\documentclass[10pt,twocolumn,letterpaper]{article}

\usepackage{iccv}              

\usepackage{graphicx}
\usepackage{booktabs}
\usepackage{amsmath}
\usepackage{amssymb}
\usepackage{siunitx}
\usepackage{hhline}
\usepackage{subcaption}
\usepackage{csquotes}
\usepackage{multirow}
\usepackage{pifont}
\usepackage[super]{nth}
\usepackage{mathtools,bbm}
\usepackage{bm}
\usepackage{textcomp}
\usepackage{xcolor}
\usepackage[utf8]{inputenc}
\usepackage{enumitem}
\usepackage{stfloats}
\usepackage[accsupp]{axessibility}  

\newcolumntype{L}[1]{>{\raggedright\let\newline\\\arraybackslash\hspace{0pt}}m{#1}}
\newcolumntype{C}[1]{>{\centering\let\newline\\\arraybackslash\hspace{0pt}}m{#1}}
\newcolumntype{R}[1]{>{\raggedleft\let\newline\\\arraybackslash\hspace{0pt}}m{#1}}

\definecolor{darkergreen}{RGB}{21, 152, 56}
\definecolor{red2}{RGB}{252, 54, 65}
\newcommand{\cmark}{\textcolor{darkergreen}{\ding{51}}}
\newcommand{\xmark}{\textcolor{red2}{\ding{55}}}

\definecolor{iccvblue}{rgb}{0.21,0.49,0.74}
\usepackage[pagebackref,breaklinks,colorlinks,allcolors=iccvblue]{hyperref}

\def\paperID{4871} 
\def\confName{ICCV}
\def\confYear{2025}

\title{PERSONA: Personalized Whole-Body 3D Avatar with Pose-Driven Deformations \\ from a Single Image}

\author{
Geonhee Sim
\hskip2.0em
Gyeongsik Moon\\
Dept. of CSE, Korea University\\
{\tt\small \{kh6362,mks0601\}@korea.ac.kr}\\
{\small \url{https://mks0601.github.io/PERSONA}}
}

\begin{document}
\maketitle

\begin{abstract}
Two major approaches exist for creating animatable human avatars. 
The first, a 3D-based approach, optimizes a NeRF- or 3DGS-based avatar from videos of a single person, achieving personalization through a disentangled identity representation. 
However, modeling pose-driven deformations, such as non-rigid cloth deformations, requires numerous pose-rich videos, which are costly and impractical to capture in daily life.
The second, a diffusion-based approach, learns pose-driven deformations from large-scale in-the-wild videos but struggles with identity preservation and pose-dependent identity entanglement.
We present PERSONA, a framework that combines the strengths of both approaches to obtain a personalized 3D human avatar with pose-driven deformations from a single image. 
PERSONA leverages a diffusion-based approach to generate pose-rich videos from the input image and optimizes a 3D avatar based on them.
To ensure high authenticity and sharp renderings across diverse poses, we introduce balanced sampling and geometry-weighted optimization. 
Balanced sampling oversamples the input image to mitigate identity shifts in diffusion-generated training videos. 
Geometry-weighted optimization prioritizes geometry constraints over image loss, preserving rendering quality in diverse poses.
\end{abstract}

\section{Introduction}

Creating an animatable human avatar is a long-standing challenge in computer vision and graphics. 
An animatable human avatar is a representation that (1) can be driven by novel whole-body poses and facial expressions and (2) can be rendered from any viewpoint.
Early approaches~\cite{peng2021neural,peng2021animatable,kwon2021neural,choi2022mononhr} relied on multi-view video data and accurately tracked 3D poses. 
While these methods achieved impressive results in controlled environments, their practical applicability was limited due to the difficulty of acquiring such data in everyday scenarios.
Recent methods have significantly lowered data requirements, allowing avatars to be built from casually captured short monocular videos~\cite{jiang2022neuman,guo2023vid2avatar,jiang2023instantavatar,kocabas2024hugs,hu2023gaussianavatar,qian20243dgs,moon2024expressive} or even a single image~\cite{xu2024magicanimate,hu2024animate,zhang2025mimicmotion,men2024mimo}, eliminating the need for multi-view recordings and precisely tracked 3D poses.

\begin{figure}[t]
\begin{center}
\includegraphics[width=\linewidth]{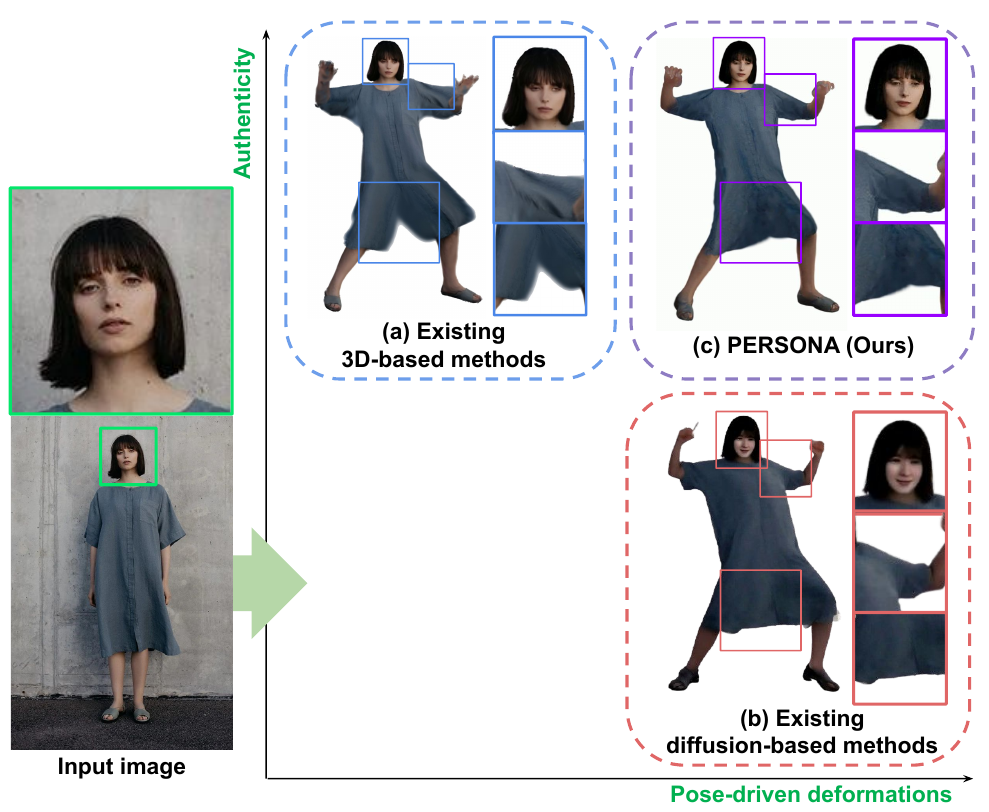}
\end{center}
\vspace*{-6mm}
\caption{
Comparison of (a) existing 3D-based method~\cite{qiu2025LHM}, (b) existing diffusion-based method~\cite{zhang2025mimicmotion}, and (c) our PERSONA.
PERSONA integrates the strengths of both approaches to achieve a personalized whole-body 3D avatar with pose-driven deformations.
}
\vspace*{-3mm}
\label{fig:intro_compare}
\end{figure}

Fig.\ref{fig:intro_compare} illustrates the two dominant approaches for creating animatable human avatars.
The first, a 3D-based approach\cite{jiang2022neuman,guo2023vid2avatar,jiang2023instantavatar,kocabas2024hugs,hu2023gaussianavatar,qian20243dgs,moon2024expressive,zhuang2024idol,qiu2025anigs,qiu2025LHM} (Fig.\ref{fig:intro_compare} (a)), combines neural rendering techniques (\emph{e.g.}, NeRF\cite{mildenhall2021nerf}, 3DGS~\cite{kerbl20233d}) with 3D human parametric models (\emph{e.g.}, SMPL~\cite{SMPL:2015}, SMPL-X~\cite{pavlakos2019expressive}).
Neural rendering captures appearance and geometry, while parametric models enable animation.
Early methods~\cite{jiang2022neuman,guo2023vid2avatar,jiang2023instantavatar,kocabas2024hugs,hu2023gaussianavatar,qian20243dgs,moon2024expressive} reconstruct avatars from short monocular videos of subjects rotating in an A-pose.
More recent approaches~\cite{zhuang2024idol,qiu2025anigs,qiu2025LHM} directly generate animatable 3D Gaussian avatars in a feed-forward manner.
This framework allows for clear disentanglement of identity and pose, enabling personalized avatar creation and faithful identity preservation.
However, as these methods animate avatars via 3D parametric models that mainly support rigid deformations, capturing pose-driven deformations such as non-rigid clothing motion requires large-scale pose-diverse datasets.
Acquiring such data per subject is costly and impractical, leading most 3D methods to rely on static or simple-pose datasets.
As a result, these avatars often lack expressiveness in handling complex, pose-dependent clothing deformations.

The second approach, a diffusion-based method (Fig.~\ref{fig:intro_compare} (b))~\cite{xu2024magicanimate,hu2024animate,zhang2025mimicmotion,men2024mimo,tu2024stableanimator,zhu2024champ}, generates animated human videos directly from a conditional 2D pose sequence using diffusion-based generative models~\cite{rombach2022high,ho2020denoising,song2020denoising,song2020improved,song2021scorebased,song2023consistency,blattmann2023stable}, without relying on neural rendering or parametric models. 
Trained on large-scale video datasets, these models effectively capture pose-driven deformations.
However, they face significant challenges in identity preservation. 
They struggle to 1) retain the identity of the person in the input image and 2) maintain identity consistency when animating avatars, resulting in limited personalization capability. 
This limitation arises because identity representation is not fully disentangled from pose, often causing pose-dependent identity variations that distort the subject’s original appearance.

We present PERSONA (Fig.~\ref{fig:intro_compare} (c)), a framework for creating personalized 3D avatars with pose-driven deformations from a single image by leveraging diffusion-generated training videos. 
This eliminates the need for extensive per-individual data capture, making the pipeline highly scalable.
PERSONA combines the strengths of 3D-based and diffusion-based approaches—3D-based methods effectively preserve identity but struggle with pose-driven non-rigid deformations in casually captured data (\emph{e.g.}, a short monocular video of a subject rotating in an A-pose). 
In contrast, diffusion-based methods capture pose-dependent deformations but lack personalization.
Our approach bridges this gap, achieving both identity preservation and pose-driven deformations in a scalable and efficient manner.

We introduce two key components to address the main challenges.
First, we propose balanced sampling to ensure high authenticity in personalization. 
Diffusion-generated training videos often fail to fully preserve identity, leading to inconsistencies across poses. 
To mitigate identity shifts, our balanced sampling oversamples the input image during avatar optimization.
In addition, we prevent baked-in artifacts such as shadows and pose-dependent geometry (\emph{e.g.}, cloth wrinkles) of the input image. 
This approach enhances identity preservation while minimizing baked-in artifacts in novel poses.

Second, we propose geometry-weighted optimization to maintain sharp renderings across diverse poses. 
Diffusion-generated videos often contain inconsistent or artifact-prone textures, and directly optimizing the avatar on such frames degrades visual quality. 
Balanced sampling alone is insufficient for preserving sharp renderings in poses different from the input image, as the pose-driven deformation modeling module differentiates between the input image and generated frames based on their pose information, causing the model to adapt to the low-quality outputs of generated frames. 
Simply detaching textures from pose-driven deformation modeling is ineffective, as image loss still encourages geometry to replicate artifacts from generated frames, leading to degraded renderings. 
To address this, geometry-weighted optimization assigns low image loss weights and high geometry loss weights. 
Since geometry (\emph{e.g.}, binary mask, depth maps, normal maps, and part segmentations) remains stable despite texture inconsistencies, it serves as a reliable constraint for non-rigid deformations. 
Additionally, omitting scale offsets in pose-driven deformation modeling prevents blurriness, significantly contributing to sharp renderings across diverse poses.

\begin{table}[t]
\footnotesize
\centering
\setlength\tabcolsep{1.0pt}
\def\arraystretch{1.1}
\begin{tabular}{L{2.5cm}|C{2.2cm}|C{1.5cm}|C{1.5cm}}
\specialrule{.1em}{.05em}{.05em}
Methods & Pose-invariant ID & Pose-driven deform. & Single img. \\ \hline
GaussianAvatar~\cite{hu2023gaussianavatar} & \cmark & \cmark & \xmark \\
ExAvatar~\cite{moon2024expressive} & \cmark & \cmark & \xmark \\ 
IDOL~\cite{zhuang2024idol} & \cmark & \xmark & \cmark \\ 
AniGS~\cite{qiu2025anigs} & \cmark & \xmark & \cmark \\ 
LHM~\cite{qiu2025LHM} & \cmark & \xmark & \cmark \\ \hline
Animate Anyone~\cite{hu2024animate} & \xmark & \cmark & \cmark \\ 
MimicMotion~\cite{zhang2025mimicmotion}  & \xmark & \cmark & \cmark \\ 
Champ~\cite{zhu2024champ} & \xmark & \cmark & \cmark \\ 
StableAnimator~\cite{tu2024stableanimator} & \xmark & \cmark & \cmark \\ \hline
\textbf{PERSONA (Ours)} & \cmark & \cmark & \cmark \\
\specialrule{.1em}{.05em}{.05em}
\end{tabular}
\vspace*{-3mm}
\caption{
Comparison of existing human avatar creation methods and the proposed PERSONA. 
The first block~\cite{hu2023gaussianavatar,moon2024expressive,zhuang2024idol,qiu2025anigs,qiu2025LHM} represents 3D-based approaches, while the second block~\cite{hu2024animate,zhang2025mimicmotion,zhu2024champ,tu2024stableanimator} corresponds to diffusion-based approaches. 
Each column indicates whether the avatar’s identity representation is pose-invariant, whether it supports pose-driven deformations (\emph{e.g.}, non-rigid cloth deformations), and whether it can be created from a single image.
}
\vspace*{-5mm}
\label{table:compare_novelty}
\end{table}

Despite the complementary strengths of 3D-based and diffusion-based methods in personalization and pose-driven deformations, few studies have effectively integrated them. 
We hope our work provides valuable insights for both research directions.
Our key contributions are as follows:
\begin{itemize}
\item We propose PERSONA, a framework for creating personalized 3D avatars with pose-driven deformations from a single image by leveraging diffusion-generated pose-rich training videos, eliminating the need for extensive per-individual video capture.
\item We introduce balanced sampling to ensure authentic identity consistency. 
It mitigates identity shifts in diffusion-generated videos while preventing baked-in artifacts such as shadows and pose-dependent geometry.
\item We propose geometry-weighted optimization, which prioritizes geometry constraints over image loss, ensuring sharp renderings across diverse poses.
\end{itemize}

\section{Related works}

Tab.~\ref{table:compare_novelty} compares existing 3D-based and diffusion-based human avatar creation methods and the proposed PERSONA.

\noindent\textbf{3D-based human avatars.}
Early works~\cite{alldieck2018video,bagautdinov2021driving,peng2021neural,peng2021animatable,zheng2023avatarrex,kwon2021neural,li2024animatable,moreau2024human,pang2024ash,kwon2024deliffas} required accurate 3D pose tracking with multi-view images.
Since accurate 3D pose tracking is rarely available in daily life, recent works focus on creating 3D avatars from casually captured monocular videos.
Jiang~\etal~\cite{jiang2022neuman} introduced NeuMan, an in-the-wild dataset with a NeRF-based baseline.
Guo~\etal~\cite{guo2023vid2avatar} proposed self-supervised scene-human decomposition, while Jiang~\etal~\cite{jiang2023instantavatar} developed a fast 3D avatar pipeline.
Kocabas~\etal~\cite{kocabas2024hugs} and Hu~\etal~\cite{hu2023gaussianavatar} leveraged 3DGS to improve representation and regression from a posed SMPL~\cite{SMPL:2015} mesh.
Liu~\etal~\cite{liu2024gea} introduced a whole-body avatar without facial expression animation, and Deng~\etal~\cite{deng2024ram} applied image-to-image translation for rendering.
Recent 3DGS-based methods further refine avatars, including Gaussian-mesh associations~\cite{shao2024splattingavatar}, hybrid surface-mesh representations~\cite{moon2024expressive}, and gradient-based optimization refinements~\cite{hu2024expressive}.
Xiu~\etal~\cite{xiu2024puzzleavatar} generated avatars from personal photo collections.
Concurrently, Zhuang~\etal~\cite{zhuang2024idol} and Qiu~\etal~\cite{qiu2025anigs} proposed single-image 3D avatar frameworks trained on static 3D scans with multi-view renderings.
However, they do not support non-rigid deformations, as capturing diverse poses and clothing behaviors is difficult, limiting the availability of suitable training data.
We address this by leveraging diffusion-generated videos, avoiding the need for extensive 3D data capture.

\noindent\textbf{Diffusion-based human avatars.}
With the success of recent generative models for image~\cite{rombach2022high,ho2020denoising,song2020denoising,song2020improved,song2021scorebased,song2023consistency} and video generation~\cite{blattmann2023stable}, dedicated generative models have been developed to animate a human from a single reference image using target 2D pose sequences.
Unlike 3D-based avatar creation methods, which rely on casually captured short monocular videos, these generative models leverage large-scale datasets to learn the prior distribution of human motion.
Xu~\etal~\cite{xu2024magicanimate} developed a video diffusion model to encode temporal dynamics, while Hu~\etal~\cite{hu2024animate} incorporated spatial attention for enhanced detail preservation.
Zhang~\etal~\cite{zhang2025mimicmotion} introduced confidence-aware pose guidance, and Men~\etal~\cite{men2024mimo} designed compact spatial encodings that account for the 3D nature of videos.
Tu~\etal~\cite{tu2024stableanimator} proposed an ID Adapter for identity preservation, and Zhu~\etal~\cite{zhu2024champ} used SMPL renderings~\cite{SMPL:2015} as a conditioning signal for human video generation.

\noindent\textbf{3D human recovery from a single image.}
3D human pose estimation methods~\cite{kanazawa2018end,kolotouros2019learning,moon2020i2l,choi2020pose2mesh,lin2021end,rong2021frankmocap,moon2022accurate} regress 3D joint angles and shape parameters of 3D human models~\cite{SMPL:2015,pavlakos2019expressive} from a single image. 
While these representations are animatable, they lack high-fidelity detail due to their reliance on simplified naked body geometry without textures. 
3D reconstruction methods~\cite{saito2019pifu,habermann2020deepcap,saito2020pifuhd,huang2020arch,zheng2021pamir,xiu2022icon,xiu2023econ,alldieck2022photorealistic,huang2024tech,shin2024canonicalfusion,kolotouros2024instant} recover posed 3D human geometry from a single image, with some approaches~\cite{saito2019pifu,huang2020arch,zheng2021pamir,alldieck2022photorealistic,huang2024tech,shin2024canonicalfusion,kolotouros2024instant} also reconstructing texture. 
Although these methods achieve high-fidelity reconstruction, their animation capability remains limited, as posed 3D scans are inherently difficult to animate~\cite{saito2021scanimate}.

\begin{figure}[t]
\begin{center}
\includegraphics[width=\linewidth]{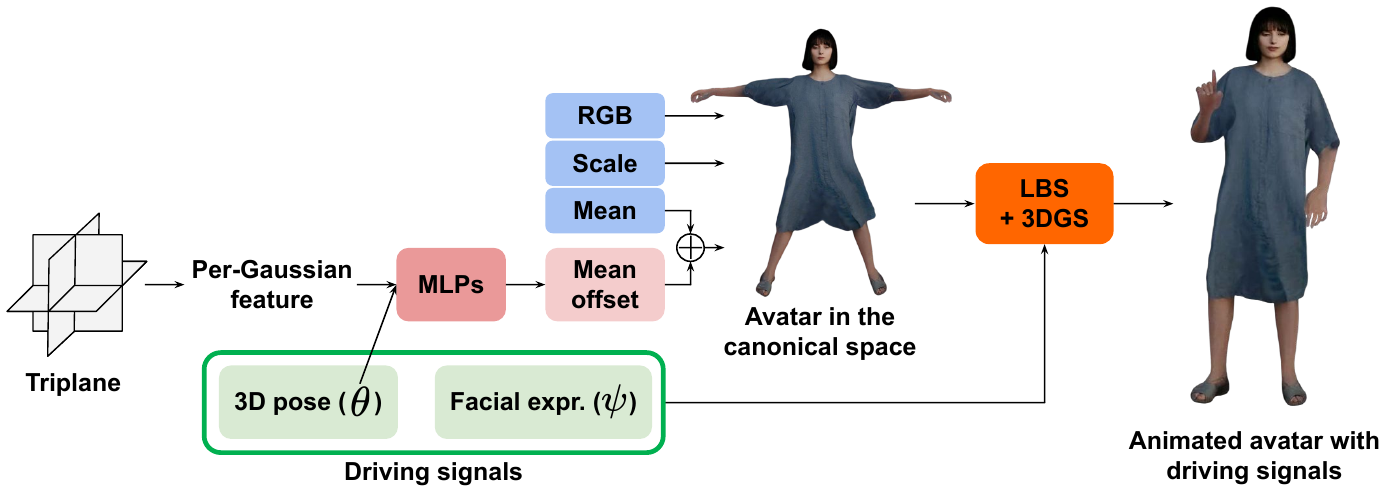}
\end{center}
\vspace*{-6mm}
\caption{
The pipeline of PERSONA.
The mean offset from MLPs is used to represent pose-driven deformations.
}
\vspace*{-3mm}
\label{fig:pipeline}
\end{figure}

\section{Pipeline of PERSONA}\label{sec:representation}

Fig.~\ref{fig:pipeline} shows the pipeline of PERSONA.

\noindent\textbf{Representation.}
We design PERSONA by combining the SMPL-X~\cite{pavlakos2019expressive} parametric model with 3D Gaussian Splatting (3DGS)~\cite{kerbl20233d,Yu2024MipSplatting}. 
SMPL-X enables whole-body animation, while 3DGS supports texture and geometry modeling along with rendering.
Following ExAvatar~\cite{moon2024expressive}, we adopt a hybrid representation of surface mesh and 3D Gaussians.
Each vertex of the SMPL-X template mesh is modeled as a 3D Gaussian, with connectivity inherited from the SMPL-X triangle topology.
To enhance generalization to novel views, we use isotropic Gaussians with constant opacity set to one.

\noindent\textbf{Architecture.}
With the optimizable Gaussian features (\emph{i.e.}, means, scales, and RGB colors), we introduce mean offsets to model pose-driven deformations. 
These offsets are predicted by multi-layer perceptrons (MLPs), which take as input the triplane features of each Gaussian point in the canonical space along with the 3D poses. 
To enhance generalization, the MLPs utilize only the 3D poses of 4-ring neighboring joints while setting non-neighboring joints to zero, following Saito~\etal~\cite{saito2021scanimate}. 
The final mean offsets, combined with facial expression-dependent vertex offsets from SMPL-X, are applied to the means in the canonical space.

\noindent\textbf{Animation and rendering.}
The 3D human avatar is constructed in a canonical space and animated using SMPL-X 3D pose $\theta$ and facial expression parameter $\psi$. 
Each body Gaussian is assigned the average skinning weight of its 16 nearest SMPL-X template vertices, while original SMPL-X weights are retained for hands and face.
The 3D Gaussians are animated using linear blend skinning (LBS), and the final rendering is performed with Mip-Splatting~\cite{Yu2024MipSplatting}.

\section{Generating training videos from an image}~\label{sec:training_videos}

\subsection{Pose-rich video generation}

\begin{figure}[t]
\begin{center}
\includegraphics[width=\linewidth]{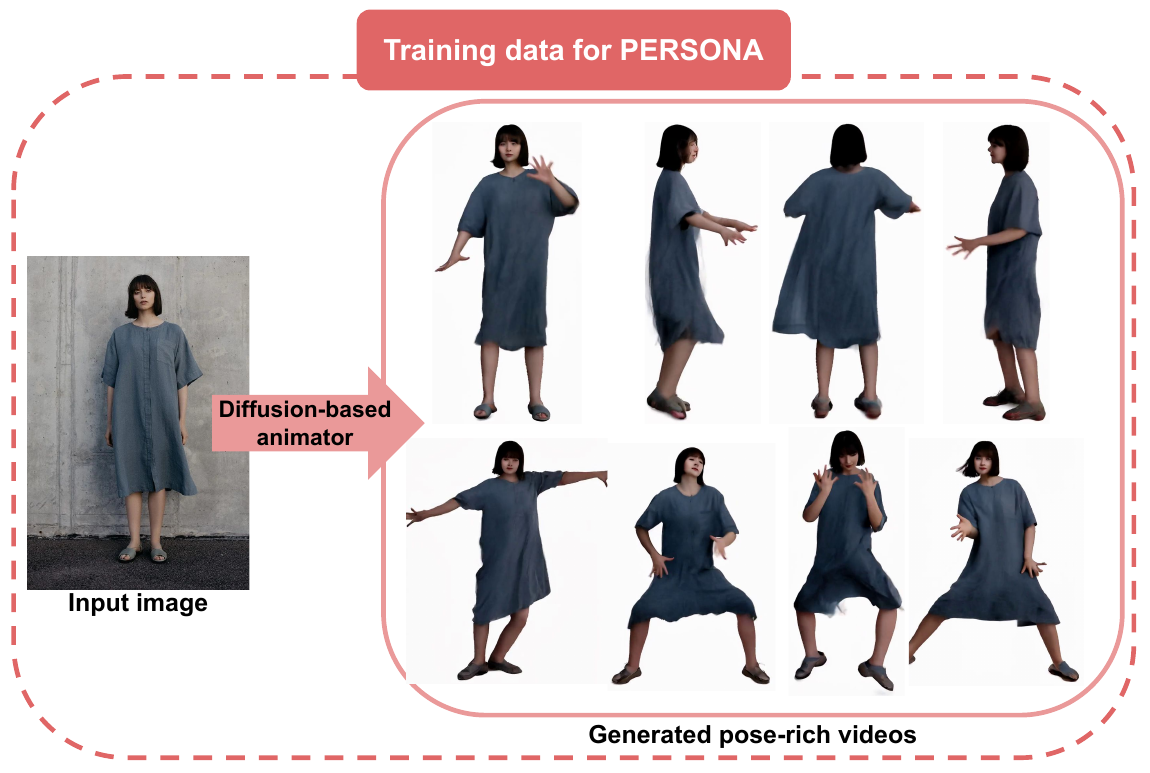}
\end{center}
\vspace*{-6mm}
\caption{
We use a diffusion-based animator~\cite{zhang2025mimicmotion} to generate pose-rich training videos from a single image. 
The input image and the generated videos together form our training set.
}
\vspace*{-5mm}
\label{fig:generate_videos}
\end{figure}

As shown in Fig.\ref{fig:generate_videos}, we generate pose-rich training videos using the diffusion-based human animation method MimicMotion~\cite{zhang2025mimicmotion}.
The generated videos and the input image are used to construct our final avatar.
These videos effectively compensate for the limited pose and deformation information available in a single image.
In particular, they reveal how clothing deforms across different poses, which is essential for modeling pose-driven (\emph{i.e.}, non-rigid) deformations.
We generate various motion types, including dance sequences with peak poses, and rotating, light punching, and kicking actions that contain milder pose variations.
We adopt MimicMotion for the training video generation as it outperforms other open-source diffusion-based methods~\cite{hu2024animate,tu2024stableanimator} for our task.

\subsection{Geometric ID-preserving video generation}~\label{subsec:geo_id_gen}

One of the key challenges in diffusion-based video generation is preserving the geometric identity of the person in the input image, such as bone lengths.
To address this, we combine identity-related SMPL-X parameters (\emph{e.g.}, shape) from the input image with target 3D poses that define the motion for animation.
These target motions include a diverse range of actions, from mild (\emph{e.g.}, rotation and light gestures) to strong (\emph{e.g.}, dancing and kicking), and are extracted in advance from public videos using the ExAvatar~\cite{moon2024expressive} fitting process.
We project the resulting SMPL-X keypoints to 2D space to create driving pose videos, which are used as input to diffusion-based animation methods~\cite{zhang2025mimicmotion,tu2024stableanimator}.
By using identity-related SMPL-X parameters from the input image, we better preserve the subject’s identity compared to prior methods~\cite{zhang2025mimicmotion,tu2024stableanimator}, which rely on aligning a small number of vertical keypoints in the 2D space.
Moreover, pre-computed 3D poses eliminate the need for the costly fitting process required by prior video-based methods~\cite{moon2024expressive,hu2023gaussianavatar}.

\begin{figure*}[t]
\begin{center}
\includegraphics[width=0.8\linewidth]{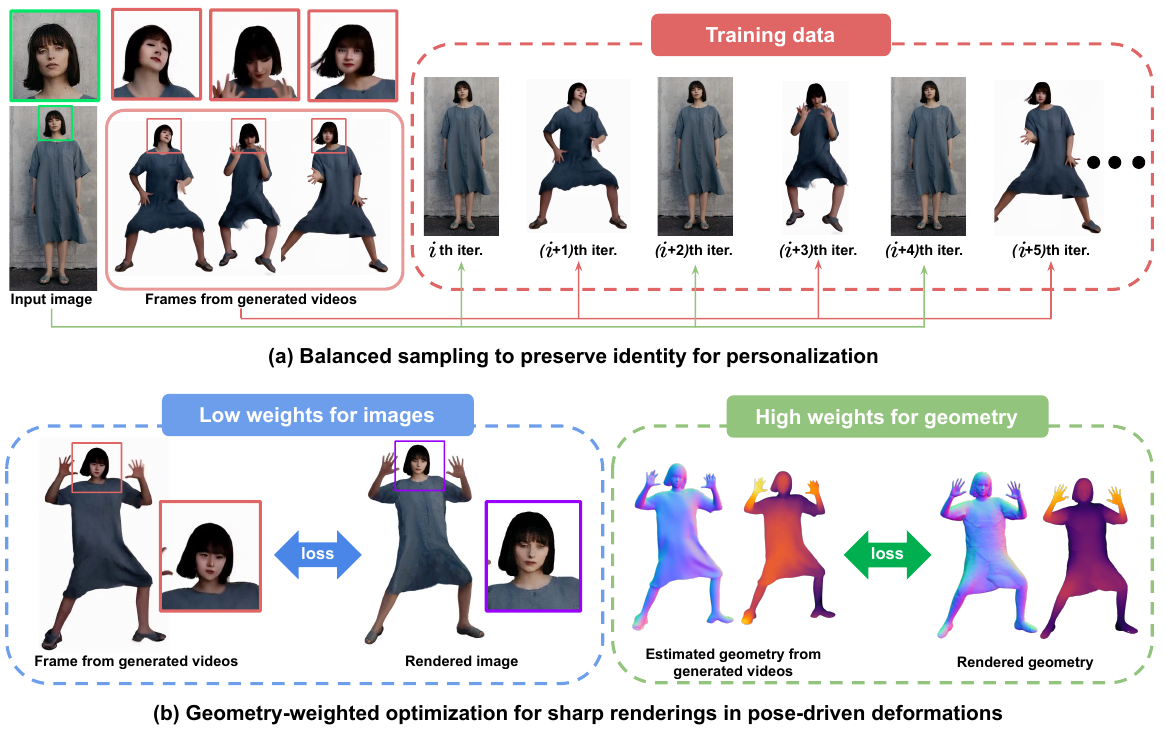}
\end{center}
\vspace*{-6mm}
\caption{
Two core components of PERSONA: balanced sampling for identity preservation and geometry-weighted optimization for sharp renderings in pose-driven deformations.
}
\vspace*{-3mm}
\label{fig:bs_and_gwo}
\end{figure*}

\section{Personalize with pose-driven deformations}

\subsection{Balanced sampling}~\label{subsec:balanced_sampling}

\noindent\textbf{Balanced sampling.}
Fig.~\ref{fig:bs_and_gwo} (a) illustrates how balanced sampling alternates between the input image and generated video frames during training, effectively oversampling the input image.
This helps prevent authenticity loss, as diffusion-generated videos often distort the subject’s identity, especially in the face region.
By using the input image more frequently, our approach maintains identity consistency in visible areas.

\begin{figure}[t]
\begin{center}
\includegraphics[width=\linewidth]{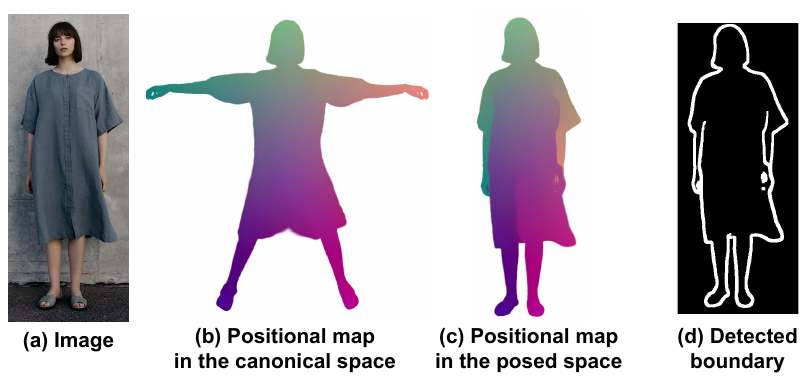}
\end{center}
\vspace*{-6mm}
\caption{
Seam boundary detection from positional maps.
We obtain (c) the positional map by applying the pose to (b), then apply a Sobel filter to (c) to detect boundaries between different body parts as well as between foreground and background.
}
\vspace*{-3mm}
\label{fig:boundary}
\end{figure}

\noindent\textbf{Reducing baked-in artifacts.}
Oversampling can introduce baked-in artifacts from the input image, including shadows in textures and seam artifacts between visible and invisible regions. 
These artifacts become embedded in the avatar’s canonical space, causing issues when animating in novel poses, as they do not naturally adapt to new viewpoints or poses.
To mitigate artifacts, we apply two strategies when supervising with the input image.

First, to reduce seam artifacts between visible and invisible regions, we identify seam boundaries by applying a Sobel filter to rendered positional maps from the canonical space, as shown in Fig.~\ref{fig:boundary}.
The positional map is obtained by encoding the normalized 3D coordinates of Gaussians in the canonical space as RGB and rendering them in the posed screen space.
Since the canonical space (\emph{i.e.}, A-pose) spatially separates body parts, abrupt changes in the rendered positional map typically indicate transitions between unrelated regions—approximating seam boundaries.
Unlike simple foreground masks, this approach can also detect internal boundaries between different body parts.
We regularize these regions using separate RGBs supervised only on generated videos, which are free from the oversampling artifacts of the input image.
Second, to avoid baked-in shadows, we use albedo images from Careaga~\etal~\cite{careaga2023intrinsic} as additional supervision, which contain minimal shading and help prevent shadow artifacts in the texture. 

\subsection{Geometry-weighted optimization}~\label{subsec:geo_weight_opt}

\noindent\textbf{Geometry-weighted optimization.}
Fig.~\ref{fig:bs_and_gwo} (b) illustrates that geometry-weighted optimization applies low image loss weights and high geometry loss weights when optimizing MLPs for pose-driven deformation modeling.
In this way, we mitigate rendering degradation caused by inconsistent and artifact-prone textures in generated frames. 
This approach enhances the robustness of the optimization pipeline against texture artifacts in diffusion-generated frames, ensuring sharper renderings in poses different from the input image.
Since per-frame pose information is used for pose-driven deformation modeling, balanced sampling alone (Sec.~\ref{subsec:balanced_sampling}) is insufficient to maintain rendering quality in poses different from the input image. 
The pose-driven deformation modeling module differentiates between the input image and generated frames based on their poses, leading to sharp renderings when the pose matches the input image, but degraded quality in novel poses as the model adapts to artifacts in generated frames. 
Simply detaching textures from pose-driven deformation modeling is ineffective, as image loss still encourages geometry to replicate blur and artifacts of the generative videos.

Geometry-weighted optimization utilizes binary masks, depth maps, normal maps, and part segmentations extracted from diffusion-generated videos using SAM~\cite{kirillov2023segment} and Sapiens~\cite{khirodkar2024sapiens} and compares them with the rendered outputs. 
Since geometry (\emph{i.e.}, binary masks, depth maps, normal maps, and part segmentations) remains stable despite variations in texture quality, it serves as a reliable foundation for modeling pose-driven deformations while preserving rendering sharpness.
The binary masks are rendered with a color value of one.
To render depth maps, the depth value of each Gaussian is treated as a color attribute. 
Normal maps are computed per Gaussian using normal vectors, leveraging the hybrid representation of ExAvatar~\cite{moon2024expressive}, and are similarly treated as colors. 
Finally, part segmentations are represented as RGB images, with colors assigned based on the Sapiens palette.

\begin{figure*}[t]
\begin{center}
\includegraphics[width=\linewidth]{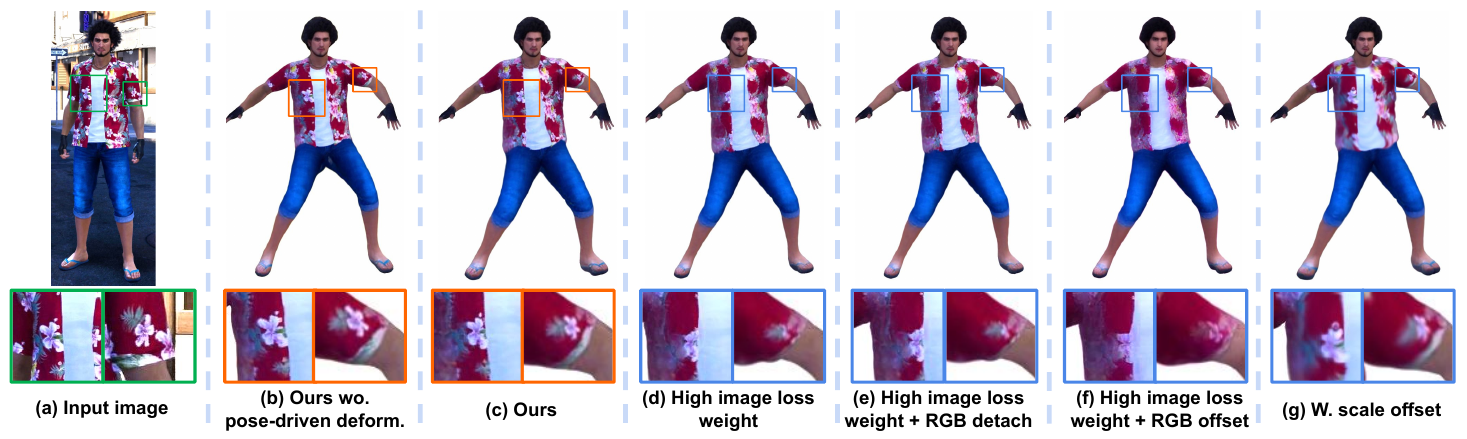}
\end{center}
\vspace*{-6mm}
\caption{
Comparison of various pose-driven deformation modeling strategies.
Our geometry-weighted optimization is essential for maintaining authentic and sharp renderings.
}
\vspace*{-3mm}
\label{fig:ablation_geo_weighted_opt}
\end{figure*}

\noindent\textbf{Preserving sharp renderings with mean offsets.}
To model pose-driven deformations, we apply only mean offsets to isotropic Gaussians.
This approach shifts each Gaussian’s position while keeping its shape and appearance fixed, allowing the avatar to deform without blurring textures.
Such position-only deformation preserves texture sharpness, similar in spirit to mesh-based animation where vertex displacements alone maintain high-frequency texture details.
In contrast, using scale offsets leads to blurry results, as Gaussians grow or shrink instead of shifting, and RGB offsets risk copying unreliable colors from generated frames.
Although mean offsets could introduce gaps, dense high-resolution Gaussians and their overlapping nature ensure seamless renderings, even during large deformations like dance motions.

\section{Optimization}

We optimize 3D Gaussian features (\emph{i.e.}, means, scales, and RGB colors), triplane, MLP weights, and per-frame SMPL-X parameters.
For supervision, we use standard image reconstruction loss functions, including $L1$, SSIM, and LPIPS~\cite{zhang2018perceptual}, along with geometry regularizers such as Laplacian regularization, following Moon~\etal~\cite{moon2024expressive}. 
In geometry-weighted optimization, we minimize the $L1$ distance between the rendered outputs and target geometry maps.
Diffusion-generated videos often produce implausible hand shapes. 
To enforce geometric plausibility, we minimize the $L1$ distance between hand masks rendered with 3DGS and those rendered with SMPL-X meshes (not Gaussian points) using a standard mesh renderer, ensuring that the hand shape closely resembles SMPL-X hands.

\begin{figure}[t]
\begin{center}
\includegraphics[width=\linewidth]{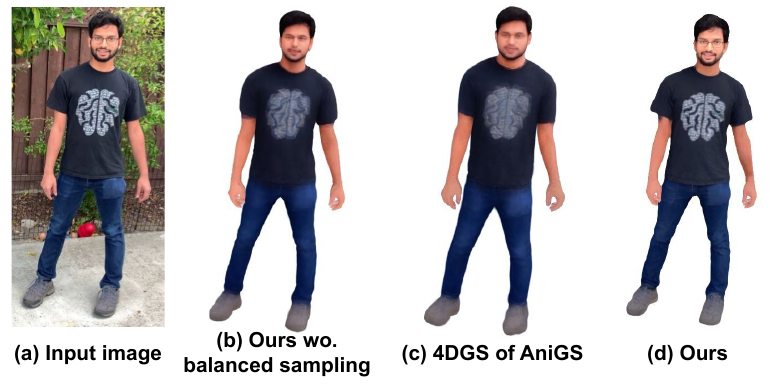}
\end{center}
\vspace*{-6mm}
\caption{
Effectiveness of our balanced sampling.
Without it, the avatar loses the identity of the input image.
The 4DGS approach of AniGS~\cite{qiu2025anigs} also fails to preserve the subject’s identity.
}
\vspace*{-5mm}
\label{fig:ablation_bs}
\end{figure}

\begin{figure}[t]
\begin{center}
\includegraphics[width=\linewidth]{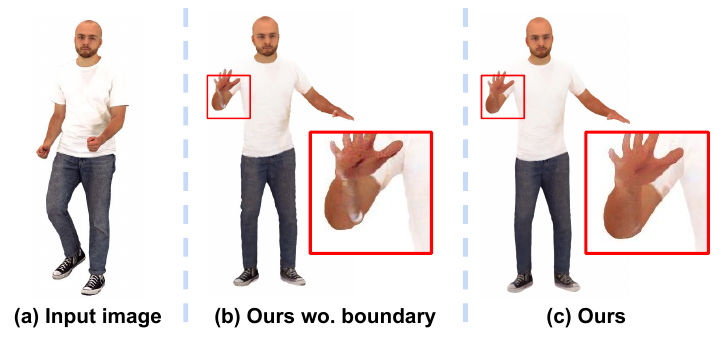}
\end{center}
\vspace*{-6mm}
\caption{
Effectiveness of our detected boundary in balanced sampling.
Without it, colors from one body part can leak into others, leading to noticeable artifacts.
}
\vspace*{-5mm}
\label{fig:ablation_bs_boundary}
\end{figure}

\section{Experiments}

\subsection{Protocol}

We conduct quantitative comparisons against state-of-the-art methods using NeuMan~\cite{jiang2022neuman} and X-Humans~\cite{shen2023x} datasets, following the evaluation protocols of previous works~\cite{moon2024expressive,shen2023x,hu2023gaussianavatar}.
We measure rendering quality with PSNR, SSIM, and LPIPS~\cite{zhang2018perceptual} metrics.
NeuMan provides short monocular videos captured in the wild, while X-Humans offers a diverse range of whole-body motions, including various body poses, hand gestures, and facial expressions. 
For qualitative comparisons, we evaluate animation capability using in-the-wild videos featuring intense dance performances, \emph{different from our training set.}

\begin{figure*}[t]
\begin{center}
\includegraphics[width=\linewidth]{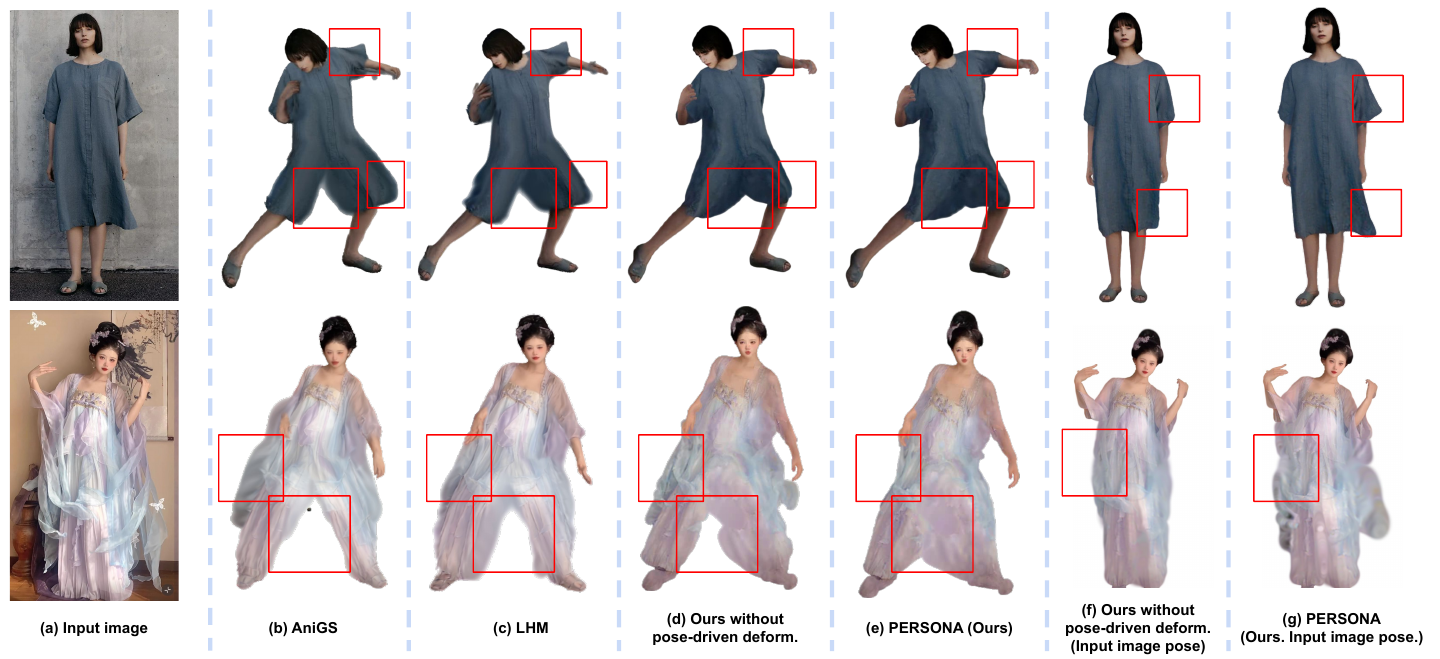}
\end{center}
\vspace*{-6mm}
\caption{
Effectiveness of our pose-driven deformations.
(b,c): Previous 3D-based methods~\cite{qiu2025anigs,qiu2025LHM} embed input-image-specific deformations (highlighted in red) into the avatar, leading to baked-in artifacts when animated to new poses.
(d-g): Our method, PERSONA, mitigates this issue by explicitly modeling pose-driven deformations.
}
\vspace*{-3mm}
\label{fig:ablation_pose_driven_deform}
\end{figure*}

\begin{figure*}[t]
\begin{center}
\includegraphics[width=\linewidth]{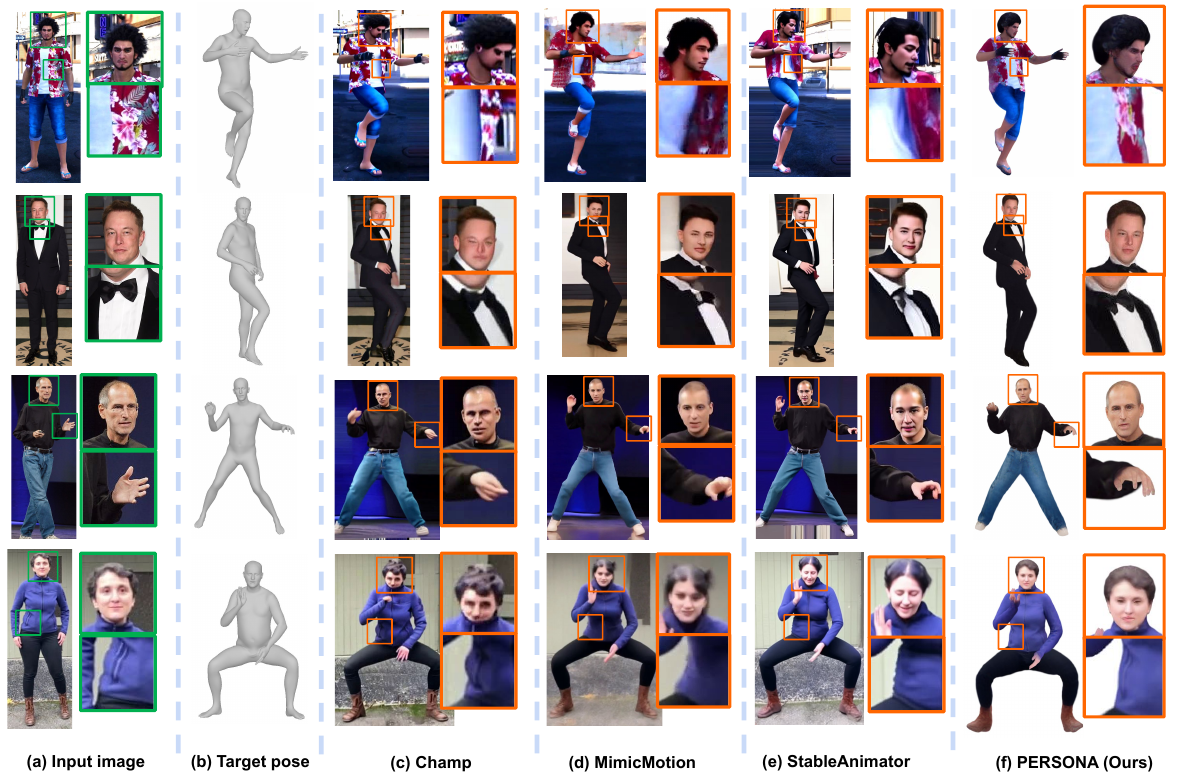}
\end{center}
\vspace*{-6mm}
\caption{
Comparison of diffusion-based state-of-the-art methods and our PERSONA.
}
\vspace*{-3mm}
\label{fig:compare_sota}
\end{figure*}

\begin{table*}[t]
\footnotesize
\centering
\setlength\tabcolsep{1.0pt}
\def\arraystretch{1.1}
\scalebox{0.95}{
\begin{tabular}{L{4.0cm}|C{1.2cm}C{1.2cm}C{1.2cm}|C{1.2cm}C{1.2cm}C{1.2cm}|C{1.2cm}C{1.2cm}C{1.2cm}}
\specialrule{.1em}{.05em}{.05em}
\multirow{2}{*}{Methods} & \multicolumn{3}{c|}{\textit{\textbf{00028}}} & \multicolumn{3}{c|}{\textit{\textbf{00034}}} & \multicolumn{3}{c}{\textit{\textbf{00087}}} \\
 & PSNR\textuparrow & SSIM\textuparrow & LPIPS\textdownarrow & PSNR\textuparrow & SSIM\textuparrow & LPIPS\textdownarrow & PSNR\textuparrow & SSIM\textuparrow & LPIPS\textdownarrow\\ \hline
 \multicolumn{10}{l}{\textbf{* Available data: 10 videos and 3D registrations from multi-view images}} \\
X-Avatar~\cite{shen2023x} & 28.57 & 0.976 & 0.026 & 28.05 & 0.965 & 0.035 & 30.89  & 0.970 & 0.030 \\
ExAvatar~\cite{moon2024expressive} & \textbf{30.58} & \textbf{0.981} & \textbf{0.018} & \textbf{28.75} & \textbf{0.966} & \textbf{0.029} & \textbf{32.01} & \textbf{0.972} & \textbf{0.025} \\ \hline
\multicolumn{10}{l}{\textbf{* Available data: a single image}} \\
ExAvatar~\cite{moon2024expressive} & 21.83  & 0.962 & 0.058 & 23.04 & 0.951  & 0.066 & 24.58 & 0.958 & 0.061 \\
TeCH~\cite{huang2024tech} &  20.28 & 0.950 & 0.059 & 22.26 & 0.938 & 0.055 & 24.74 & 0.945 & 0.052 \\
SiTH~\cite{ho2024sith} & 15.00 & 0.934 & 0.110 & 18.28 & 0.935 & 0.098 & 21.65 & 0.950 & 0.087 \\
Champ~\cite{zhu2024champ} &  18.96 & 0.953 & 0.070 & 20.17 & 0.949 & 0.082 & 25.19 & 0.960 & 0.057  \\
MimicMotion~\cite{zhang2025mimicmotion} &  21.02 & 0.959 & 0.053 & 23.49 & 0.953 & 0.051 & 27.67 & 0.963 & 0.038  \\
StableAnimator~\cite{tu2024stableanimator} &  21.66 & 0.962 & 0.050 & 23.53 & 0.952 & 0.051 & 28.04 & 0.963 & 0.039  \\
AniGS~\cite{qiu2025anigs} & 23.96 & 0.965 & \textbf{0.040} & 24.80 & 0.955 & 0.052 & 27.46 & 0.964 & 0.043 \\
\textbf{Ours wo. pose-driven deform.} & 23.15 & 0.968  & 0.049 &  26.32 & 0.960 &  0.047 &  29.25 & 0.966  &  0.038 \\ 
\textbf{PERSONA (Ours)} & \textbf{24.76} &  \textbf{0.972} & \textbf{0.040} &  \textbf{27.60} & \textbf{0.963} & \textbf{0.042} &  \textbf{29.79} &  \textbf{0.968} &  \textbf{0.035} \\ 
\specialrule{.1em}{.05em}{.05em}
\end{tabular}
}
\vspace*{-3mm}
\caption{
Comparisons with previous works on the test set of X-Humans~\cite{shen2023x}.
}
\label{table:compare_sota_xhumans}
\vspace*{-3mm}
\end{table*}

\subsection{Ablation studies}

\noindent\textbf{Balanced sampling.}
Fig.~\ref{fig:ablation_bs} demonstrates the importance of balanced sampling in preserving authenticity.
Without it, as shown in Fig.~\ref{fig:ablation_bs} (b), the input image is underused during training, resulting in an avatar with a different identity and blurry textures due to inconsistencies in the generated frames.
Fig.~\ref{fig:ablation_bs} (c) shows that 4DGS-based method~\cite{qiu2025anigs} still suffers from identity loss and blurry renderings due to the severe imbalance between the input image and the generated frames.
Their 4DGS treats inconsistencies across frames as a temporal sequence and trains MLPs with spatio-temporal features to differentiate the input image from generated frames.
Fig.~\ref{fig:ablation_bs_boundary} shows that our boundary detection is necessary to prevent color leaking between different body parts.

\noindent\textbf{Geometry-weighted optimization.}
Fig.~\ref{fig:ablation_geo_weighted_opt} highlights the importance of geometry-weighted optimization in preserving authentic and sharp renderings for pose-driven deformations. 
Without it, inconsistent textures in generated frames cause identity shifts and blurriness. 
Fig.~\ref{fig:ablation_geo_weighted_opt} (c) demonstrates effective pose-driven deformation, where raising the arms causes the clothing to move further away from the waist, reflecting a natural interaction between the body and the fabric. 
In contrast, Fig.~\ref{fig:ablation_geo_weighted_opt} (b) shows unnatural adherence, with the clothing remaining tightly fitted to the body despite the raised arms due to the lack of deformation modeling.
Fig.~\ref{fig:ablation_geo_weighted_opt} (d) and (e) illustrate how high image loss weights degrade visual quality, forcing geometry to replicate texture inconsistencies and resulting in blurry renderings, even when RGB is detached. 
Finally, Fig.~\ref{fig:ablation_geo_weighted_opt} (f) and (g) validate our choice to use only mean offsets, as scale offsets lead to excessive blurring, as discussed in Sec.~\ref{subsec:geo_weight_opt}. 
All variants are evaluated under the same balanced sampling and geometry loss settings for a fair comparison.

\noindent\textbf{Pose-driven deformations.}
Fig.\ref{fig:ablation_pose_driven_deform} illustrates the effectiveness of our pose-driven deformations.
Existing methods\cite{qiu2025anigs,qiu2025LHM} embed deformations present in the input image, which become baked-in artifacts when the avatar is animated to novel poses.
In the first row, for example, the left arm and thigh should fall naturally due to gravity, but remain unnaturally bent.
In the second row, the right side of the body similarly fails to respond naturally to gravity.
In both cases, long skirts are often misinterpreted as pants, resulting in incorrect deformation behavior.
In contrast, PERSONA explicitly models pose-driven deformations, allowing avatars to respond naturally to novel poses without inheriting input-specific artifacts.
This leads to significantly improved visual fidelity under diverse poses.

\begin{table}[t]
\footnotesize
\centering
\setlength\tabcolsep{1.0pt}
\def\arraystretch{1.1}
\scalebox{0.9}{
\begin{tabular}{L{4.2cm}|C{1.2cm}C{1.2cm}C{1.2cm}}
\specialrule{.1em}{.05em}{.05em}
Methods & PSNR\textuparrow & SSIM\textuparrow & LPIPS\textdownarrow \\ \hline
\multicolumn{4}{l}{\textbf{* Available data: a video}} \\
HumanNeRF~\cite{weng2022humannerf} & 27.06 & 0.967 & 0.019 \\
InstantAvatar~\cite{jiang2023instantavatar} & 28.47 & 0.972 & 0.028 \\
NeuMan~\cite{jiang2022neuman} & 29.32 & 0.972 & 0.014 \\  
Vid2Avatar~\cite{guo2023vid2avatar} & 30.70 & 0.980 & 0.014 \\
GaussianAvatar~\cite{hu2023gaussianavatar} & 29.94 & 0.980 & 0.012 \\
3DGS-Avatar~\cite{qian20243dgs} & 28.99 & 0.974 & 0.016 \\
ExAvatar~\cite{moon2024expressive} & \textbf{34.80} & \textbf{0.984} & \textbf{0.009} \\  \hline
\multicolumn{4}{l}{\textbf{* Available data: a single image}} \\
ExAvatar~\cite{moon2024expressive} & 24.95 & 0.963 & 0.031 \\
TeCH~\cite{huang2024tech} & 22.82 & 0.953 & 0.039 \\
SiTH~\cite{ho2024sith} & 23.96 & 0.957 & 0.031 \\
Champ~\cite{zhu2024champ} & 27.27 & 0.968 & \textbf{0.021} \\
MimicMotion~\cite{zhang2025mimicmotion} & 26.12 & 0.970 & 0.029 \\
StableAnimator~\cite{tu2024stableanimator} & 26.58 & 0.968 & 0.025 \\
AniGS~\cite{qiu2025anigs} & 28.27 & 0.969 & 0.027 \\
LHM~\cite{qiu2025LHM} & 26.22 & 0.967 & 0.025 \\
\textbf{Ours wo. pose-driven deform.} & 28.02 &  0.972 & 0.025 \\ 
\textbf{PERSONA (Ours)} & \textbf{29.20} & \textbf{0.974} & \textbf{0.021} \\ 
\specialrule{.1em}{.05em}{.05em}
\end{tabular}
}
\vspace*{-3mm}
\caption{
Comparisons of previous works on the test set of NeuMan~\cite{jiang2022neuman}.
}
\label{table:compare_sota_neuman}
\vspace*{-3mm}
\end{table}

\subsection{Comparisons to state-of-the-art methods}

Fig.\ref{fig:ablation_pose_driven_deform}, Fig.\ref{fig:compare_sota}, Tab.\ref{table:compare_sota_xhumans}, and Tab.\ref{table:compare_sota_neuman} compare PERSONA with state-of-the-art methods~\cite{qiu2025anigs,qiu2025LHM,zhu2024champ,zhang2025mimicmotion,tu2024stableanimator}.
Fig.\ref{fig:ablation_pose_driven_deform} shows that compared to existing 3D-based approaches~\cite{qiu2025anigs,qiu2025LHM}, PERSONA captures natural pose-driven clothing deformations more effectively, thanks to our geometry-weighted optimization.
In addition, Fig.\ref{fig:compare_sota} shows that compared to diffusion-based approaches~\cite{zhu2024champ,zhang2025mimicmotion,tu2024stableanimator}, PERSONA better preserves identity, accurately maintaining facial features and clothing patterns through balanced sampling.
Tab.\ref{table:compare_sota_xhumans} and Tab.\ref{table:compare_sota_neuman} show that PERSONA outperforms all single-image-based methods, demonstrating the necessity and effectiveness of pose-driven deformations.
All comparisons exclude background pixels and use official implementations.

\section{Conclusion}

We introduce PERSONA, a framework that creates personalized 3D avatars with pose-driven deformations from a single image using diffusion-generated training videos. 
By combining 3D- and diffusion-based approaches, PERSONA ensures identity preservation and natural deformations. 
To address authenticity loss and rendering artifacts, we propose balanced sampling and geometry-weighted optimization. 
Our results show that PERSONA outperforms existing methods, providing a scalable solution for high-quality avatar creation.

\clearpage

\begin{center}
\textbf{\large Supplementary Material \textit{for} \\ \vspace{2mm}
\large{``PERSONA: Personalized Whole-Body 3D Avatar \\with Pose-Driven Deformations from a Single Image"}}
\end{center}

\setcounter{section}{0}
\setcounter{table}{0}
\setcounter{figure}{0}
\renewcommand{\thesection}{S\arabic{section}}   
\renewcommand{\thetable}{S\arabic{table}}   
\renewcommand{\thefigure}{S\arabic{figure}}

In this supplementary material, we provide more experiments, discussions, and other details that could not be included in the main text due to the lack of pages.
The contents are summarized below:
\begin{itemize}[nosep]
    \item Sec.~\ref{sec:compare_sota_suppl}: More comparisons to state-of-the-art methods.
    \item Sec.~\ref{sec:avatar_canon_space}: Rendered avatars in the canonical space.
    \item Sec.~\ref{sec:ablation_suppl}: More ablation studies.
    \item Sec.~\ref{sec:limitations_suppl}: Limitations of the proposed PERSONA.
\end{itemize}

\section{Comparisons to state-of-the-art methods}~\label{sec:compare_sota_suppl}

\noindent\textbf{Running time comparison.}
Tab.~\ref{table:compare_sota_running_time} further highlights that PERSONA achieves real-time rendering speeds, whereas existing diffusion-based methods suffer from slow inference.
All running times were measured under the same hardware setup using a single RTX A6000.

\begin{figure}[t]
\begin{center}
\includegraphics[width=0.8\linewidth]{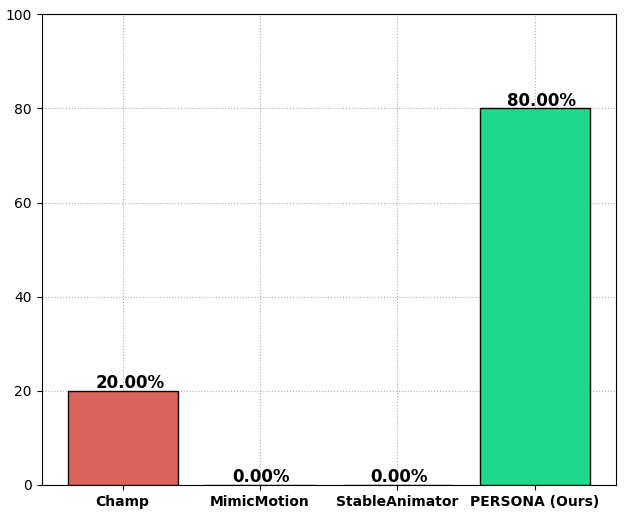}
\end{center}
\vspace*{-6mm}
\caption{
User preference study results from 40 participants.
}
\vspace*{-5mm}
\label{fig:user_study}
\end{figure}

\begin{figure}[t]
\begin{center}
\includegraphics[width=0.9\linewidth]{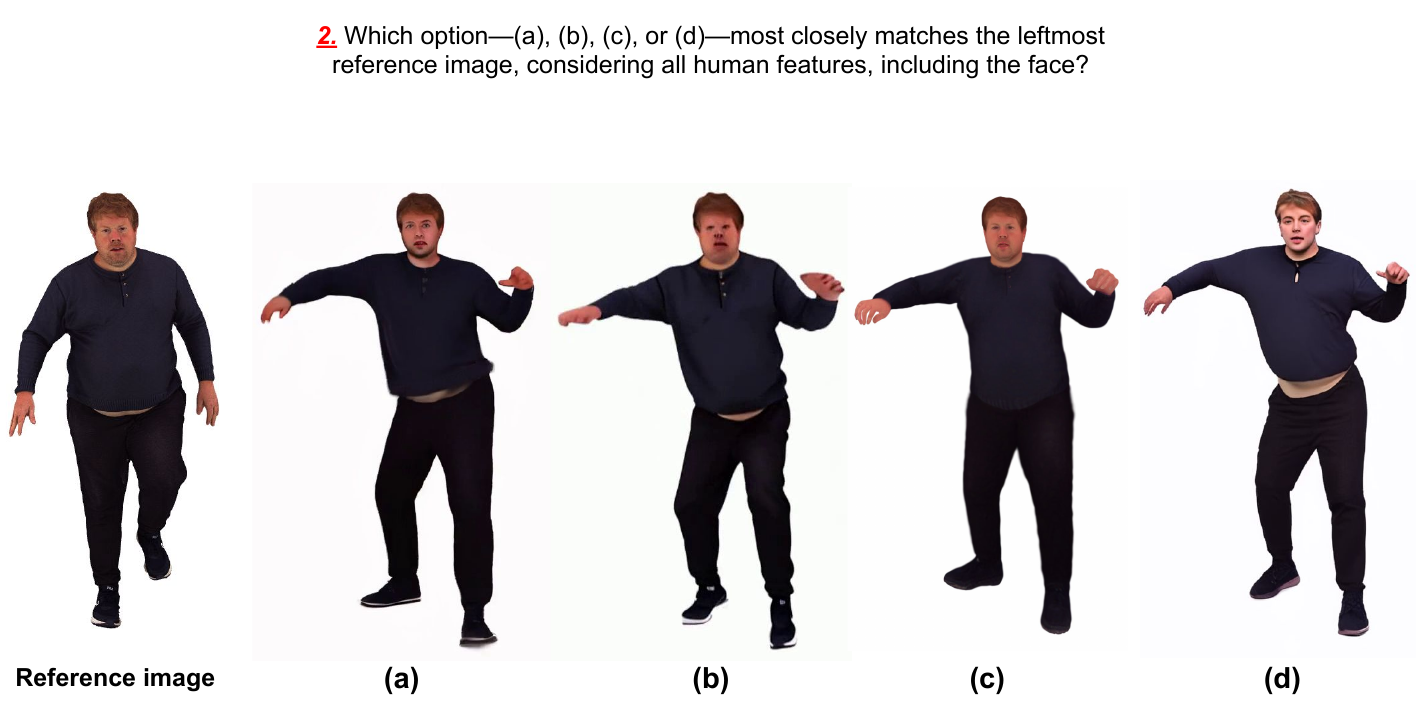}
\end{center}
\vspace*{-7mm}
\caption{
An example of our user study.
}
\vspace*{-4mm}
\label{fig:user_study_example}
\end{figure}

\noindent\textbf{User study.}
Fig.\ref{fig:user_study} presents results from our user study, where participants strongly preferred our approach over existing diffusion-based methods.
We conducted the study with 40 participants, each answering 10 questions in which they selected the image that best matched the input single image.
The compared methods included Champ~\cite{zhu2024champ}, MimicMotion~\cite{zhang2025mimicmotion}, StableAnimator~\cite{tu2024stableanimator}, and our PERSONA.
Fig.~\ref{fig:user_study_example} provides an example from the study, with (a), (b), (c), and (d) corresponding to MimicMotion~\cite{zhang2025mimicmotion}, Champ~\cite{zhu2024champ}, our PERSONA, and StableAnimator~\cite{tu2024stableanimator}, respectively.

\begin{figure*}[t]
\begin{center}
\includegraphics[width=\linewidth]{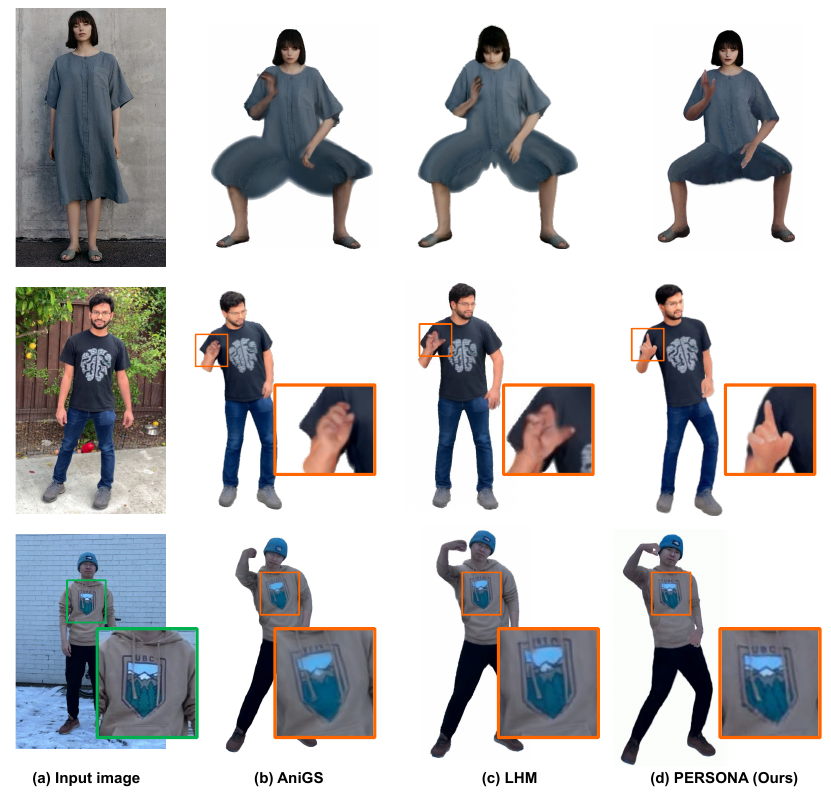}
\end{center}
\vspace*{-6mm}
\caption{
Comparison of state-of-the-art 3D-based methods~\cite{qiu2025anigs,qiu2025LHM} and our PERSONA.
}
\vspace*{-5mm}
\label{fig:compare_sota_3d_suppl}
\end{figure*}

\begin{figure*}[t]
\begin{center}
\includegraphics[width=\linewidth]{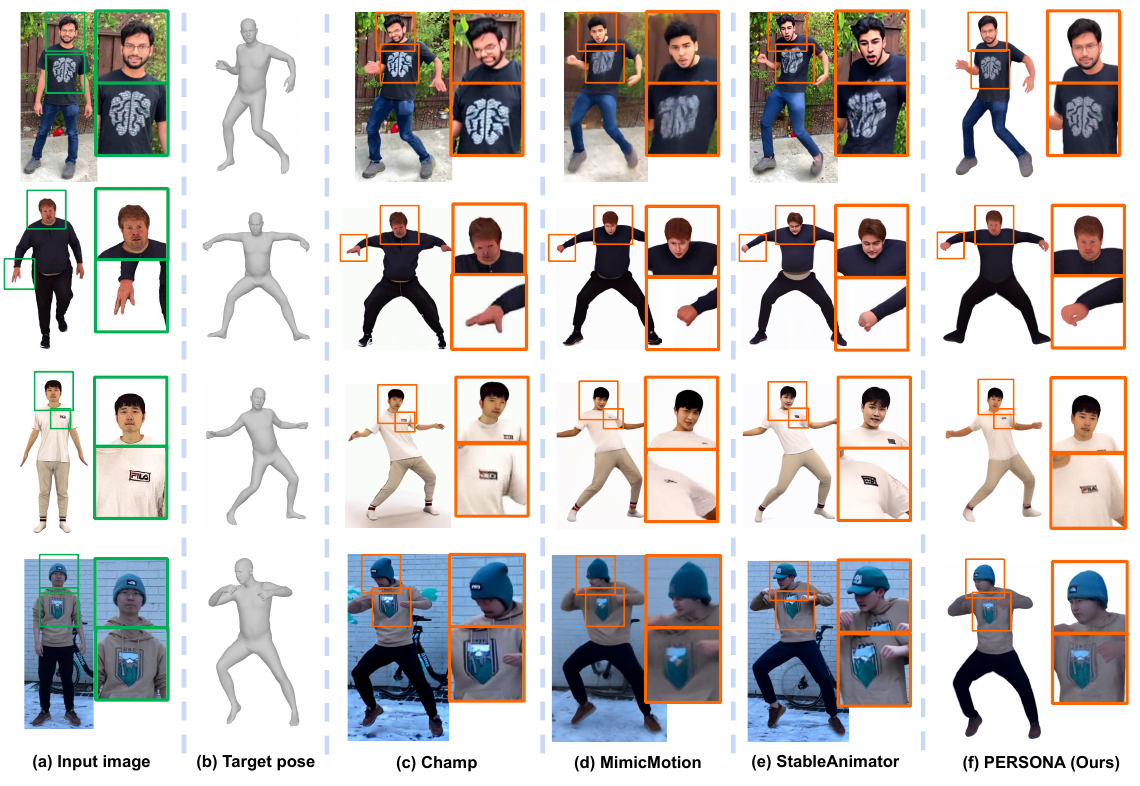}
\end{center}
\vspace*{-6mm}
\caption{
Comparison of state-of-the-art diffusion-based methods~\cite{zhu2024champ,zhang2025mimicmotion,tu2024stableanimator} and our PERSONA.
}
\vspace*{-5mm}
\label{fig:compare_sota_gen_suppl}
\end{figure*}

\noindent\textbf{Qualitative comparisons.}
Fig.\ref{fig:compare_sota_3d_suppl} compares our PERSONA with 3D-based state-of-the-art methods\cite{qiu2025anigs,qiu2025LHM}.
PERSONA achieves more accurate pose-driven deformations with more stable and consistent renderings.
Fig.\ref{fig:compare_sota_gen_suppl} compares PERSONA with diffusion-based methods\cite{zhu2024champ,zhang2025mimicmotion,tu2024stableanimator}, where our method better preserves the subject’s identity from the input image, resulting in more authentic avatars while still accurately modeling pose-driven deformations.

 \newpage

\begin{table}[t]
\footnotesize
\centering
\setlength\tabcolsep{1.0pt}
\def\arraystretch{1.1}
\scalebox{1.0}{
\begin{tabular}{L{2.5cm}|C{3.0cm}}
\specialrule{.1em}{.05em}{.05em}
Methods & Frames per second \\ \hline
Champ~\cite{zhu2024champ} &  0.88 \\
MimicMotion~\cite{zhang2025mimicmotion} & 0.36 \\
StableAnimator~\cite{tu2024stableanimator} & 0.24 \\
\textbf{PERSONA (Ours)} & \textbf{25.56} \\ 
\specialrule{.1em}{.05em}{.05em}
\end{tabular}
}
\vspace*{-3mm}
\caption{
Frames per second comparisons of various human animation methods.
}
\label{table:compare_sota_running_time}
\vspace*{-3mm}
\end{table}

\clearpage

\begin{figure*}[b]
\begin{center}
\includegraphics[width=0.85\linewidth]{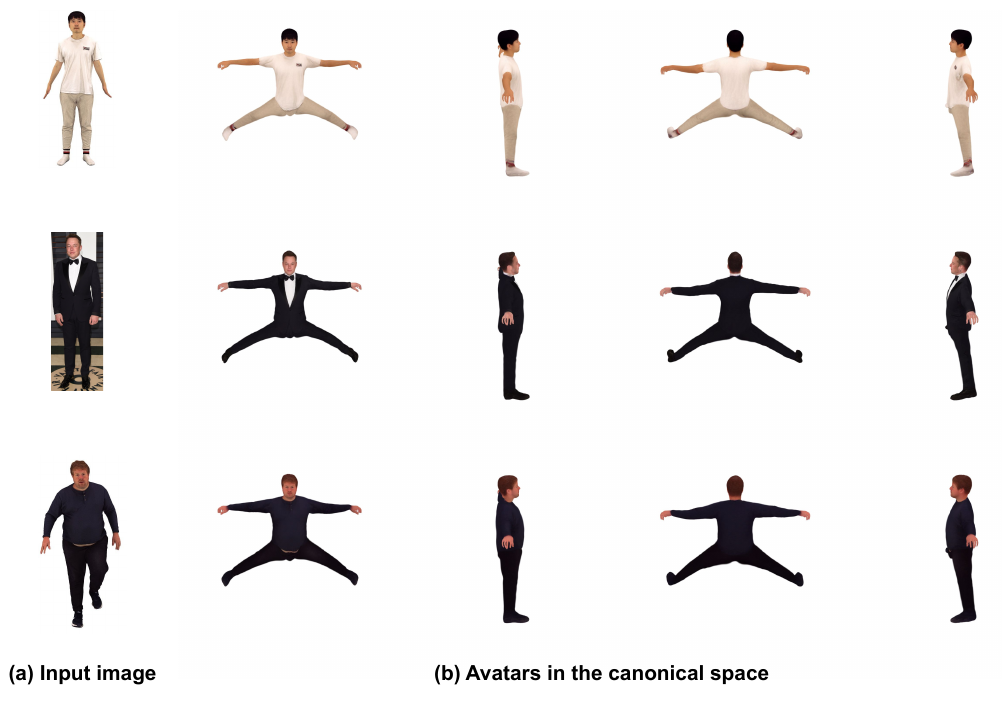}
\end{center}
\vspace*{-6mm}
\caption{
The input image and rendered avatars in the canonical space from multiple viewpoints.
}
\vspace*{-3mm}
\label{fig:neutral_pose_1}
\end{figure*}

\section{Avatars in canonical space}~\label{sec:avatar_canon_space}

\begin{figure*}[t]
\begin{center}
\includegraphics[width=0.8\linewidth]{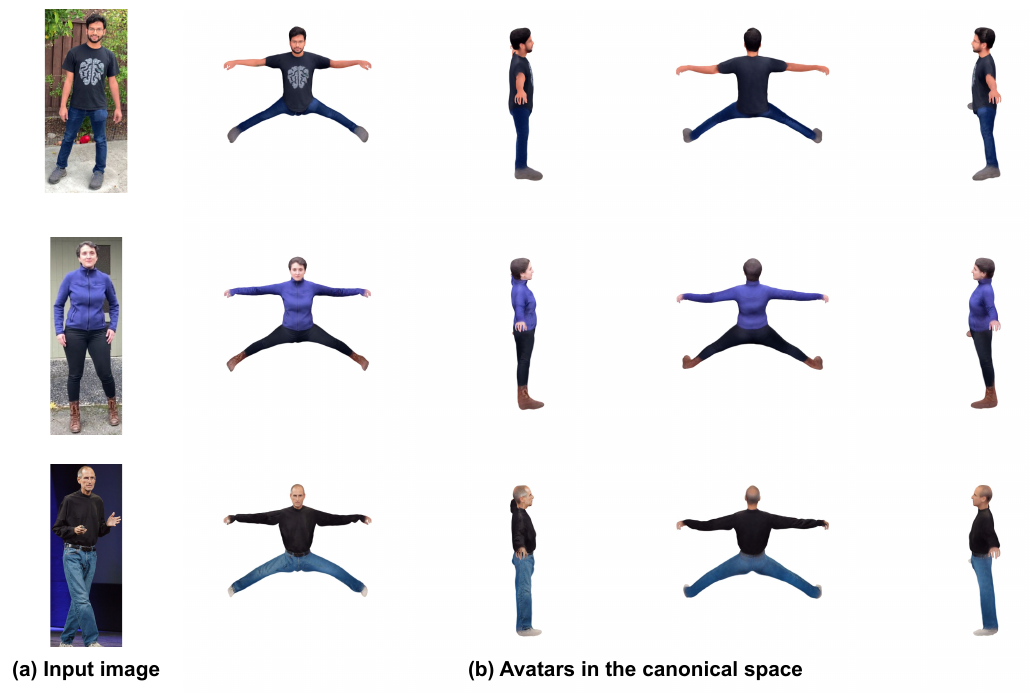}
\end{center}
\vspace*{-6mm}
\caption{
The input image and rendered avatars in the canonical space from multiple viewpoints.
}
\vspace*{-3mm}
\label{fig:neutral_pose_2}
\end{figure*}

\begin{figure*}[t]
\begin{center}
\includegraphics[width=0.8\linewidth]{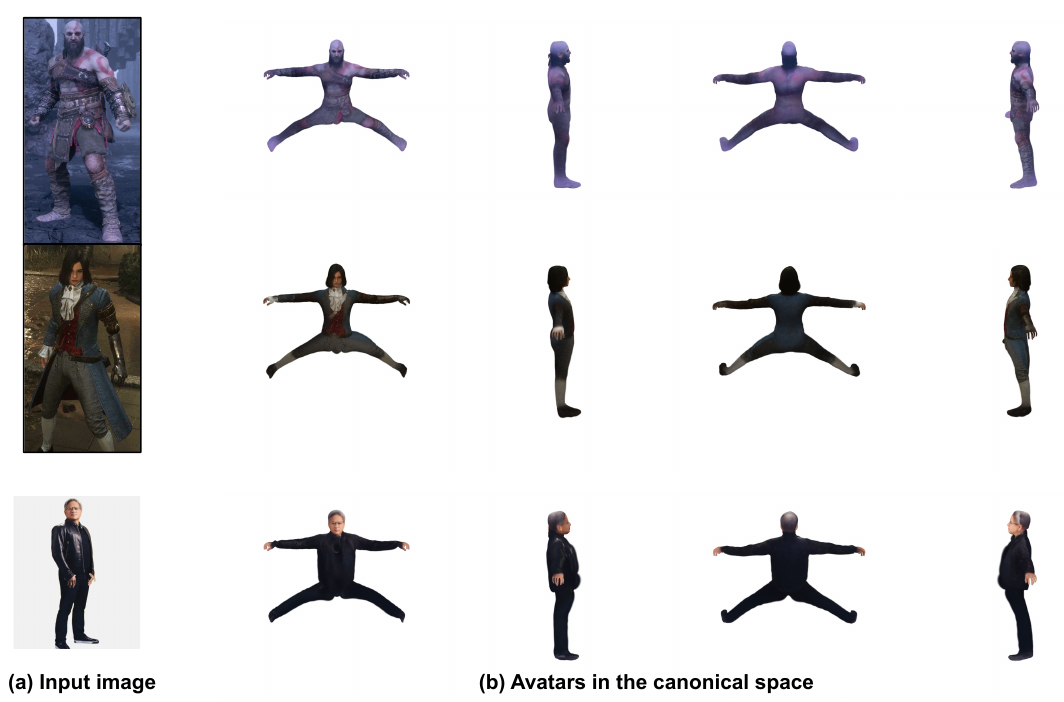}
\end{center}
\vspace*{-6mm}
\caption{
The input image and rendered avatars in the canonical space from multiple viewpoints.
}
\vspace*{-3mm}
\label{fig:neutral_pose_3}
\end{figure*}

Fig.~\ref{fig:neutral_pose_1}, ~\ref{fig:neutral_pose_2}, and ~\ref{fig:neutral_pose_3} showcase various avatars created from a single input image.
These avatars are rendered in canonical space without applying our pose-driven deformations.
Despite being constructed from just a single image, the avatars achieve high-quality renderings from multiple viewpoints, including fully invisible regions, without noticeable artifacts.
These results highlight the effectiveness of our avatar creation pipeline.

\clearpage

\begin{figure}[t]
\begin{center}
\includegraphics[width=\linewidth]{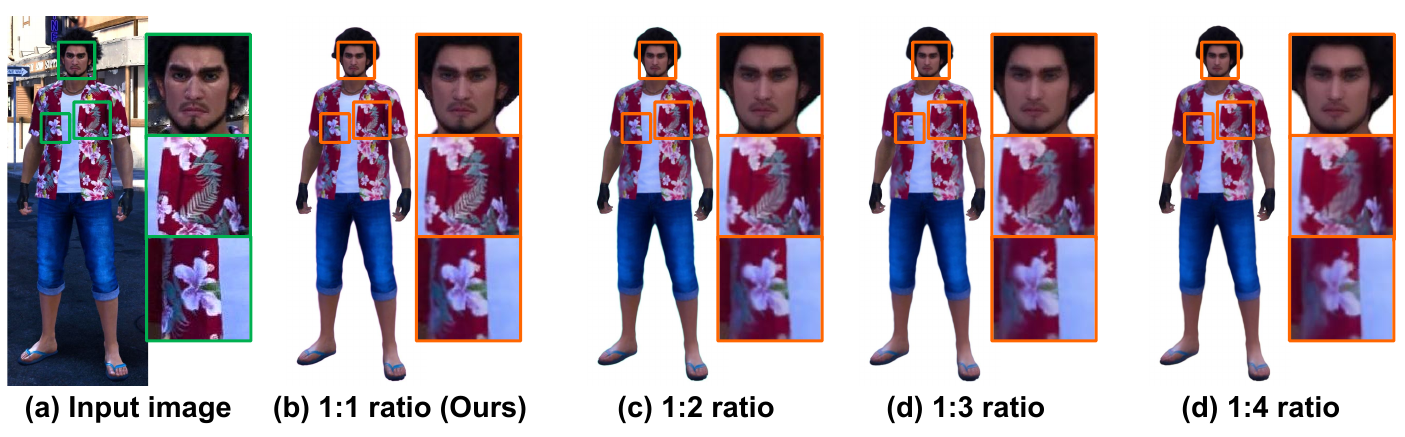}
\end{center}
\vspace*{-6mm}
\caption{
Rendering comparisons with different input image-to-generated frame ratios in balanced sampling.
For a clearer comparison, avatars are rendered using the viewpoint and pose of the input image.
}
\vspace*{-3mm}
\label{fig:ablation_bs_ratio}
\end{figure}

\begin{table}[t]
\footnotesize
\centering
\setlength\tabcolsep{1.0pt}
\def\arraystretch{1.1}
\scalebox{1.0}{
\begin{tabular}{C{1.5cm}C{1.5cm}|C{1.2cm}C{1.2cm}C{1.2cm}}
\specialrule{.1em}{.05em}{.05em}
Geo. weight & Img. weight & PSNR\textuparrow & SSIM\textuparrow & LPIPS\textdownarrow \\ \hline
  1 & 1 & 28.18 & 0.969 & 0.030 \\ 
  \textbf{1} & \textbf{0.1} & \textbf{29.20} & \textbf{0.974} & \textbf{0.021} \\ 
  1 & 0.01 & 29.00 & 0.970 & 0.023 \\ 
\specialrule{.1em}{.05em}{.05em}
\end{tabular}
}
\vspace*{-3mm}
\caption{
Effect of loss weights in our geometry-weighted optimization on the NeuMan test set.
The second row (in bold) is ours.
}
\vspace*{-3mm}
\label{table:ablation_geo_loss_weight}
\end{table}

\begin{table}[t]
\footnotesize
\centering
\setlength\tabcolsep{1.0pt}
\def\arraystretch{1.1}
\scalebox{1.0}{
\begin{tabular}{L{4.0cm}|C{1.2cm}C{1.2cm}C{1.2cm}}
\specialrule{.1em}{.05em}{.05em}
Settings & Mask\textuparrow & Depth\textdownarrow & Normal\textdownarrow \\ \hline
Wo. pose-driven deform. & 88.60 & 47.17 & 22.07 \\
\textbf{W. pose-driven deform. (Ours)} & \textbf{90.06} & \textbf{46.13} & \textbf{21.73} \\
\specialrule{.1em}{.05em}{.05em}
\end{tabular}
}
\vspace*{-3mm}
\caption{
Effectiveness of our pose-driven deformations on the X-Humans~\cite{shen2023x} test set. 
Units for mask, depth, and normal are \%, mm, and degrees, respectively.
}
\label{table:ablation_pose_driven_deform}
\vspace*{-3mm}
\end{table}

\begin{table}[t]
\footnotesize
\centering
\setlength\tabcolsep{1.0pt}
\def\arraystretch{1.1}
\scalebox{1.0}{
\begin{tabular}{L{3.0cm}|C{1.2cm}C{1.2cm}C{1.2cm}}
\specialrule{.1em}{.05em}{.05em}
Generator & PSNR\textuparrow & SSIM\textuparrow & LPIPS\textdownarrow \\ \hline
Champ & 29.13 & 0.972 & \textbf{0.019} \\ 
StableAnimator & 28.98 & 0.970 & 0.024 \\ 
\textbf{MimicMotion (Ours)} & \textbf{29.20} & \textbf{0.974} & 0.021 \\ 
\specialrule{.1em}{.05em}{.05em}
\end{tabular}
}
\vspace*{-3mm}
\caption{
Effect of different training video generators on the NeuMan test set.
}
\vspace*{-3mm}
\label{table:ablation_various_generators}
\end{table}

\begin{table}[t]
\footnotesize
\centering
\setlength\tabcolsep{1.0pt}
\def\arraystretch{1.1}
\scalebox{1.0}{
\begin{tabular}{L{2.5cm}|C{1.2cm}C{1.2cm}C{1.2cm}}
\specialrule{.1em}{.05em}{.05em}
Sapiens models & PSNR\textuparrow & SSIM\textuparrow & LPIPS\textdownarrow \\ \hline
 0.3B (Smallest one) & 28.98 & 0.971 & 0.023 \\ 
 \textbf{1B (Ours)} & \textbf{29.20} & \textbf{0.974} & \textbf{0.021} \\ 
\specialrule{.1em}{.05em}{.05em}
\end{tabular}
}
\vspace*{-3mm}
\caption{
Effect of different geometry estimators on the NeuMan test set.
}
\vspace*{-3mm}
\label{table:ablation_various_sapiens}
\end{table}

\section{Ablation studies}~\label{sec:ablation_suppl}

\noindent\textbf{Balanced sampling.}
Fig.~\ref{fig:ablation_bs_ratio} demonstrates the effectiveness of our 1:1 ratio between the input image and generated frames in balanced sampling.
Reducing the use of the input image leads to a loss of authenticity and sharpness in the rendering, which is expected due to the inconsistent textures in the generated frames.

\noindent\textbf{Loss weights for geometry-weighted optimization.}
Tab.~\ref{table:ablation_geo_loss_weight} shows that using a high image loss weight (first row) significantly degrades rendering quality.
This issue is mitigated by lowering the image loss weight (second row).
However, further reducing it slightly harms rendering quality (third row), indicating the need for a balanced trade-off.

\noindent\textbf{Pose-driven deformations.}
Tab.~\ref{table:ablation_pose_driven_deform} demonstrates that our pose-driven deformation not only improves photometric metrics (as shown in Tab.~\ref{table:compare_sota_xhumans} and ~\ref{table:compare_sota_neuman}) but also enhances geometry quality. 
Mask, depth, and normal metrics are measured as intersection-over-union, $L1$ distance between rendered and ground truth depth maps after aligning global translation, and the angular difference between rendered and ground truth normal maps, respectively. 

\noindent\textbf{Variants in pre-processing stages.}
Tab.~\ref{table:ablation_various_generators} and Tab.~\ref{table:ablation_various_sapiens} show how different training video generators (Sec.~\ref{sec:training_videos}) and geometry estimators (Sec.~\ref{subsec:geo_weight_opt}) affect the final rendering quality.
As shown in the tables, the choice of generator or the use of lighter geometry estimators has only a marginal impact on rendering quality.
In particular, since we use enough number of generated frames (approximately 1K) for optimizing PERSONA, the geometric estimation errors from lighter models such as Sapiens~\cite{khirodkar2024sapiens} do not significantly degrade the final output.

\begin{figure}[t]
\begin{center}
\includegraphics[width=\linewidth]{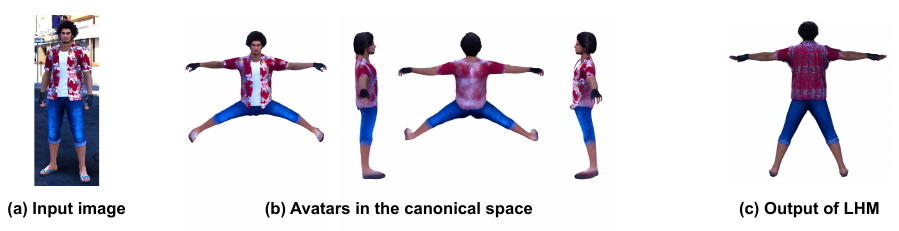}
\end{center}
\vspace*{-6mm}
\caption{
Limitation of PERSONA.
Due to texture inconsistencies of generated frames, used to train our PERSONA, complex patterns in invisible regions are challenging to render sharply.
Even very recent feed-forward method~\cite{qiu2025LHM} fail to generate plausible textures.
}
\vspace*{-5mm}
\label{fig:neutral_pose_limitation}
\end{figure}

\section{Limitations}~\label{sec:limitations_suppl}

\noindent\textbf{Lack of dynamics.}
Despite its ability to represent pose-driven deformations, PERSONA cannot capture motion-dependent dynamics, which rely on velocity and acceleration. 
These dynamics are crucial for modeling complex deformations in loose-fitting clothing and hair. 
While we attempted to incorporate velocity and acceleration as additional inputs, our 3D avatar representation lacks separate layers for garments and hair, leading to unsatisfactory results. 
We believe that designing separate layers for garments and hair could be an interesting direction for future research.

\noindent\textbf{Lack of fine-grained cloth wrinkles.}
Additionally, PERSONA struggles to capture fine, pose-dependent wrinkles in clothing, likely due to the lack of 3D consistency in diffusion-generated videos, which hinders accurate geometry and texture tracking and results in oversmoothed surfaces. 

\noindent\textbf{Blurry rendering for complex patterns in invisible regions.}
Fig.~\ref{fig:neutral_pose_limitation} illustrates that our pipeline struggles to achieve sharp renderings in invisible regions when complex patterns are present.
While our method produces plausible geometry and textures for these areas, as seen in Fig.~\ref{fig:neutral_pose_1} and Fig.~\ref{fig:neutral_pose_2}, intricate patterns remain difficult to render sharply due to inconsistencies in the generated frames used to train PERSONA.
We observe that even recent feed-forward methods~\cite{qiu2025LHM} fail to generate plausible textures.
We believe this limitation could be addressed by incorporating more advanced image or video generative models.

\noindent\textbf{Lack of relighting capability.}
Lastly, omitting RGB offsets in pose-driven deformation modeling prevents our method from handling relighting effects, such as natural shadows and reflections in novel environments. 
Addressing these challenges remains an avenue for future work.

\noindent\textbf{Long pre-processing time.}
Generating training videos with diffusion-based animators requires significant pre-processing time due to their slow inference speed. 
It takes approximately one hour to generate training videos, whereas avatar training itself additionally takes 30 minutes. 
Exploring strategies to optimize data generation for a more efficient avatar creation pipeline presents an interesting direction for future research.

\section*{Acknowledgments}
This work was partly supported by the ICT Creative Consilience Program through the Institute of Information \& Communications Technology Planning \& Evaluation (IITP) grant funded by the Korea government (MSIT) (IITP-2025-RS-2020-II201819).
It was also supported by IITP grant funded by the MSIT (RS-2025-02653113, High-Performance Research AI Computing Infrastructure Support at the 2 PFLOPS Scale).
Additional support was provided by Korea University research grants.
It was also supported by the Artificial Intelligence Industrial Convergence Cluster Development Project funded by the MSIT and Gwangju Metropolitan City.
This research was supported by the Culture, Sports and Tourism R\&D Program through the Korea Creative Content Agency grant funded by the Ministry of Culture, Sports and Tourism in 2025 (Project Name: Development of AI-based image expansion and service technology for high-resolution (8K/16K) service of performance contents, Project Number: RS-2024-00395886, Contribution Rate: 20\%).
This work was supported by the Technology Innovation Program (RS-2025-02653087, Development of a Motion Data Collection System and Dynamic Persona Modeling Technology) funded by the Ministry of Trade, Industry \& Energy (MOTIE, Korea).
This work was supported by the IITP grant funded by the MSIT (No. RS-2025-25441838, Development of a human foundation model for human-centric universal artificial intelligence and training of personnel).

\clearpage

{
\small
\bibliographystyle{ieeenat_fullname}
\bibliography{main}
}

\end{document}